\documentclass{article}
\usepackage{ijcai19}

\usepackage{times}
\usepackage{soul}
\usepackage{url}
\usepackage[hidelinks]{hyperref}
\usepackage[utf8]{inputenc}
\usepackage[small]{caption}
\usepackage{graphicx}
\usepackage{amsmath}
\usepackage{booktabs}
\usepackage{algorithm}
\usepackage{algorithmic}
\usepackage{microtype}
\usepackage{color}
\usepackage{amssymb}
\usepackage{subcaption}
\usepackage{xspace}

\graphicspath{{./}{figs/}}
\DeclareMathOperator*{\argmax}{arg\,max}

\newcommand{\nth}{\textsuperscript{th}\xspace}
\newcommand{\codeurl}{https://github.com/llan-ml/tesp}

\title{Meta Reinforcement Learning with Task Embedding and Shared Policy}
\author{
Lin Lan$^1$\footnote{This work was primarily done during the author’s internship at Huawei Noah’s Ark Lab.}\and
Zhenguo Li$^2$\and
Xiaohong Guan$^{1,3,4}$\And
Pinghui Wang$^{3,1}$\footnote{Corresponding Author.}\\
\affiliations
$^1$MOE NSKEY Lab, Xi’an Jiaotong University, China\\
$^2$Huawei Noah’s Ark Lab\\
$^3$Shenzhen Research School, Xi’an Jiaotong University, China\\
$^4$Department of Automation and NLIST Lab, Tsinghua University, China\\
\emails
llan@sei.xjtu.edu.cn,
li.zhenguo@huawei.com,
\{xhguan, phwang\}@mail.xjtu.edu.cn
}

\begin{document}

\maketitle

\begin{abstract}
Despite significant progress, deep reinforcement learning (RL) suffers from data-inefficiency and limited generalization.
Recent efforts apply meta-learning to learn a meta-learner from a set of RL tasks such that a novel but related task could be solved quickly.
Though specific in some ways,
different tasks in meta-RL are generally similar at a high level.
However, most meta-RL methods do not explicitly and adequately model the specific and shared information
among different tasks,
which limits their ability to learn training tasks and to generalize to novel tasks.
In this paper,
we propose to capture the shared information on the one hand
and meta-learn how to quickly abstract the specific information about a task on the other hand.
Methodologically, we train an SGD meta-learner to quickly optimize a task encoder for each task,
which generates a task embedding based on past experience.
Meanwhile, we learn a policy
which is shared across all tasks and conditioned on task embeddings.
Empirical results\footnote{Code available at \url{\codeurl}.}
on four simulated tasks demonstrate that
our method has better learning capacity on both training and novel tasks
and attains up to 3 to 4 times higher returns compared to baselines.

\end{abstract}

\section{Introduction}
Reinforcement learning (RL) aims to guide an agent to take actions in an environment
such that the cumulative reward is maximized~\cite{sutton1998reinforcement}. Recently,
deep RL has achieved great progress in applications such as AlphaGo~\cite{silver2016mastering}, playing Atari games~\cite{mnih2013playing}, and robotic control~\cite{levine2016end} by using deep neural networks.
However, existing RL methods suffer from data-inefficiency and limited generalization,
since they learn each task from scratch without reusing past experience,
even though these tasks are quite similar.
Recent progress in meta-learning
has shown its power to solve few-shot classification
problems~\cite{finn2017model,snell2017prototypical},
which can learn a model for a novel few-shot classification task in just a few iterations.
In this paper, we further investigate to apply meta-learning to RL domains (called meta-RL).

Basically, meta-RL consists of two modules:
policy and meta-learner.
The former defines the network structure
mapping observations to actions,
and the latter is applied to optimize a policy (i.e., learn a set of parameters)
for each task.
The objective of meta-RL is to train a meta-learner from a set of RL tasks,
which can quickly optimize a policy to solve a novel but related task.
In effect, meta-RL explores how to solve a family of tasks
rather than a single task as in conventional RL.

A major limitation of most meta-RL methods
(discussed thoroughly in \S~\ref{sec:related})
is that they
do not explicitly and adequately model the individuality and the commonness of tasks,
which has proven to play an important role in the literature of multi-task learning~\cite{ruder2017overview,ma2018modeling}
and should be likewise applicable to meta-RL.
Take the case of locomotion tasks,
where an agent needs to move to different target locations for different tasks.
The nature of this type of tasks (i.e., the commonness) is
the way to control the agent to move from one location to another,
and different tasks are mainly distinguished by the corresponding target locations (i.e., the individuality).
Humans have a similar mechanism to solve such decision making problems.
Imagine that when we want to walk to some different places,
we do not need to modify the method we walk, but modify the destinations we want to go.
Therefore, we hypothesize that a more principled approach for meta-RL is
to characterize the commonness of all tasks on the one hand
and meta-learn how to quickly abstract the individuality of each task
on the other hand.

\begin{figure}[t]
\centering
\begin{subfigure}[b]{0.32\textwidth}
\includegraphics[width=\textwidth]{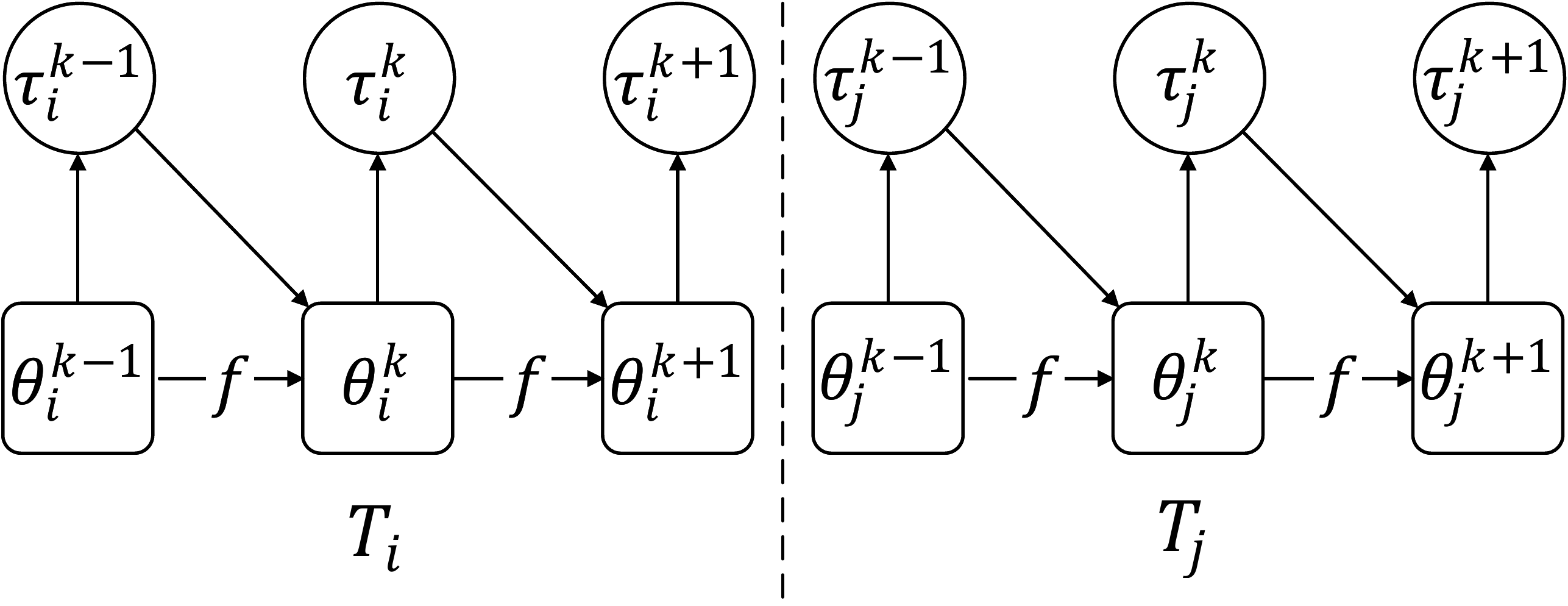}
\caption{MAML~\cite{finn2017model}.}
\label{fig:cg_maml}
\end{subfigure}
\begin{subfigure}[b]{0.48\textwidth}
\vspace{8pt}
\includegraphics[width=\textwidth]{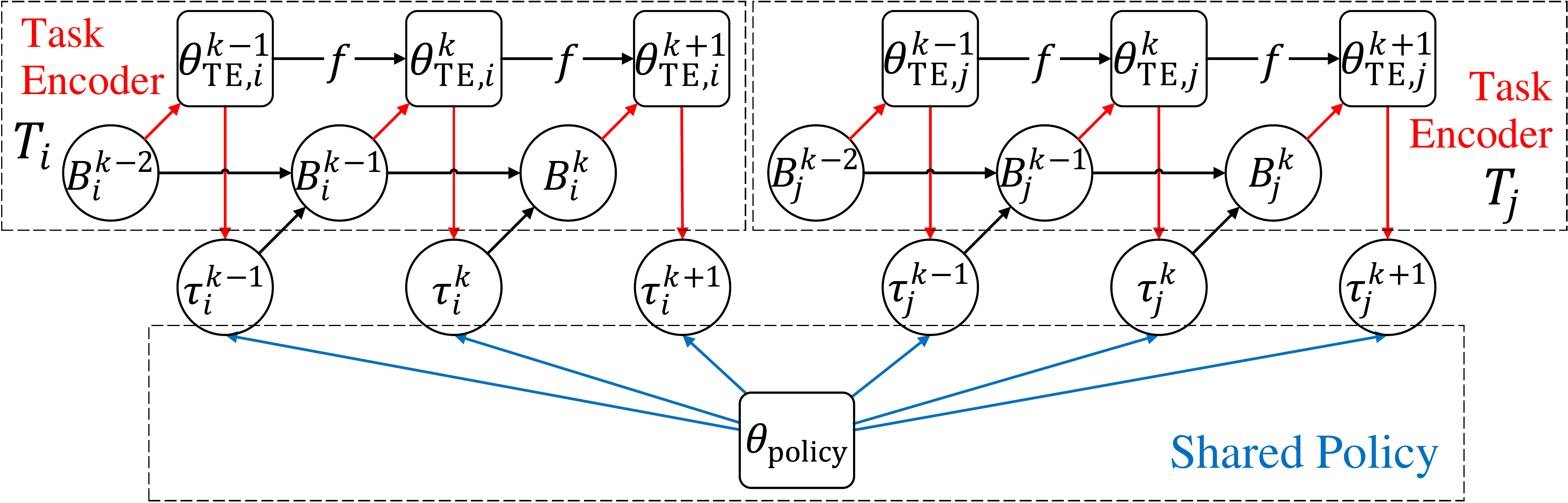}
\caption{TESP.}
\label{fig:cg_tasp}
\end{subfigure}
\caption{
Computation graphs of
MAML
and TESP.
(\subref{fig:cg_maml})
The meta-learner $f$ of MAML optimizes a policy $\theta_i$ for each task $T_i$ via
standard policy gradient using sampled episodes $\tau_i$.
(\subref{fig:cg_tasp})
TESP trains a meta-learner $f$ and a shared policy.
The meta-learner optimizes a task encoder (TE) $\theta_{\text{TE},i}$ for each task
$T_i$ based on previously sampled episodes stored in the episode buffer $B_i$.
The policy $\theta_{\text{policy}}$ is shared across tasks
and accomplishes a task based on observations and the output of the corresponding task encoder.
}
\label{fig:cg}
\end{figure}

Based on the above motivation, we introduce a new component into the current meta-RL framework,
named task encoder, and develop a new meta-RL method,
which achieves better performance on both training and novel tasks
with Task Encoder adaptation and Shared Policy, namely TESP.
Figure~\ref{fig:cg} illustrates the computation graph of TESP.
Instead of training a meta-learner that directly optimizes a policy for each task
(i.e., policy adaptation),
TESP trains a meta-learner
to quickly optimize a task encoder for each task (i.e., task encoder adaptation).
The task-specific encoder generates a latent task embedding
based on
past experience (i.e.,
previously explored episodes)
stored in an episode buffer. At the same time,
TESP trains a shared policy across different tasks.
The shared policy is conditioned on task embeddings,
which allows to accomplish different tasks based on the corresponding task embeddings
with the same set of policy parameters.

The main idea behind TESP is that
we apply the meta-learner to quickly abstract the individuality of each task
via task encoder adaptation,
and the shared policy characterizes the commonness of a family of tasks.
We evaluate our method on a variety of simulated tasks:
locomotion with a wheeled agent,
locomotion with a quadrupedal agent,
2-link reacher, and 4-link reacher.
Empirical results
show that our method has better learning capacity on both training and novel tasks.

\section{Related Work}
\label{sec:related}

The works most related to ours are MAML~\cite{finn2017model} and Meta-SGD~\cite{li2017meta}.
Specifically,
MAML trains a parametrized stochastic gradient descent (SGD) optimizer as the meta-learner,
which is expected to have a good network initialization such that
different tasks can be learned quickly
with vanilla policy gradient (VPG)~\cite{williams1992simple}.
Meta-SGD further extends MAML by introducing adaptive per-parameter learning rates.
To a certain extent, the initialization and adaptive learning rates encode the commonness of different tasks.
However, the task-specific information (i.e., the individuality) can only be implicitly obtained through subsequent policy gradient update,
which is sparse and delayed, and not effective enough for exploration in RL.
In contrast,
we introduce a meta-learned task encoder to explicitly abstract the individuality of each task
represented by a task embedding.
For each task, the task embedding is then fed into a policy network at each timestep,
which leads to dense and immediate task-specific guidance.
On the other hand,
we encoder
the commonness of a kind of tasks
into a shared policy,
rather than the parameters of the SGD optimizer.

Another related work is MAESN~\cite{gupta2018meta},
which additionally meta-learns a latent variable to capture the task-specific information
based on MAML. The variable is fed into a policy network and held constant
over the duration of episodes as well.
However, we observe that simply adapting a single variable is not enough
to represent a task
in our experiments
(conducted in a more challenging way than~\cite{gupta2018meta}).
Meanwhile,
there are
some hierarchical RL (HRL) works that involve optimizing a latent variable and have a similar network architecture to TESP.
For example, \cite{florensa2017stochastic}
pre-learns a policy conditioned on skills represented by a latent variable $z$,
and uses the pre-learned policy conditioned on task-specific skills to learn different tasks.
The task-specific skills are obtained by training extra neural networks with $z$ as input.
The latent variables learned by the above works can also be regarded as
task embeddings, which, to some extent, are learned in a transductive-like way.
The key difference is that
our method tries to induce a general function to acquire task embeddings
from episodes that have been experienced in the past,
which should be more generalizable to novel tasks.
On the other hand,
conventional HRL methods usually cannot
learn novel tasks quickly (e.g., in $3$ iterations).

MLSH~\cite{frans2017meta} also introduces the concept of ``shared policy'', which learns a set of shared policies across all tasks
and meta-learns a master policy to choose different policies in different time periods
for each task.
We think TESP and MLSH are developed from different perspectives and should be complementary to each other.
In particular, TESP can be further extended with a set of shared conditional policies, which we leave as future work.
On the other hand, the master policy of MLSH makes decisions based on observations,
which could be further improved by conditioning on a task embedding
output by a (meta-learned) task encoder.

Another line of work is to use a recurrent architecture to
act as the meta-learner.
For instance,
\cite{duan2016rl}
meta-learns a recurrent neural network (RNN)
which learns a task by updating the hidden state via the rollout and preserving the hidden state across episode boundaries.
\cite{mishra2017meta} further designs a more complex recurrent architecture
based on temporal convolutions and soft attention.
These methods encode the task individuality into the internal state of the meta-learner (e.g., the hidden state of RNN).
However,
depending on the feed-forward calculation to learn a task
seems to lead to completely overfitting to the distribution of training tasks and fail to
learn novel tasks sampled from a different distribution as shown in~\cite{houthooft2018evolved}.
Some prior works~\cite{kang2018transferable,tirinzoni2018transfer} show that MAML also suffers from this problem to some extent.

Other recent works mainly explore meta-RL
from different perspectives about what to meta-learn,
such as the exploration ability~\cite{stadie2018some},
the replay buffer for training DDPG~\cite{lillicrap2015continuous,xu2018learning},
non-stationary dynamics~\cite{al2017continuous},
factors of temporal difference~\cite{xu2018meta},
the loss function~\cite{houthooft2018evolved},
the environment model for model-based RL~\cite{clavera2018learning},
and
the reward functions in the context of
unsupervised learning and inverse RL respectively~\cite{gupta2018unsupervised,xu2018learninginverse}.
Interested readers could refer to the reference citations for more details.

\section{Preliminaries}
In this section, we first formally define the problem of meta-RL,
and then introduce a typical meta-learning (or meta-RL) method,
MAML~\cite{finn2017model},
for consistency.

\subsection{Meta-RL}
In meta-RL, we consider a set of tasks $D$,
of which each is a Markov decision process (MDP).
We denote each task by $T_i=(S,A,H,P_i,r_i)$,
where $S$ is the state space
\footnote{We use the terms \emph{state} and \emph{observation} interchangeably throughout this paper.},
$A$ is the action space, $H$ is the horizon (i.e., the maximum length of an episode),
$P_i\colon S\times A\times S\to \mathbb{R}_{\geqslant 0}$ is the transition probability distribution,
and $r_i\colon S\times A\to \mathbb{R}$ is the reward function.
Tasks have the same state space, action space, and horizon, while
the transition probabilities and reward functions differ across tasks.

Given the state $s_{i,t}\in S$ perceived from the environment at time $t$ for task $T_i$,
a policy $\pi_{\theta_i}\colon S\times A\to \mathbb{R}_{\geqslant 0}$, parametrized by $\theta_i$, predicts a distribution of actions, from which
an action $a_{i,t}\sim \pi_{\theta_i}(a_{i,t}|s_{i,t})$ is sampled.
The agent moves to the next state $s_{i,t+1}\sim P_i(s_{i,t+1}|s_{i,t},a_{i,t})$,
and receives an immediate reward $r_{i,t}=r_i(s_{i,t},a_{i,t})$.
As the agent repeatedly interacts with the environment,
an episode $\tau_i=(s_{i,0},a_{i,0},r_{i,0},...,s_{i,t},a_{i,t},r_{i,t},...)$ is collected,
and it stops when the termination condition is reached or the length of $\tau_i$ is $H$.
We denote by $\tau_i\sim P_{i}(\tau_i|\theta_i)$ sampling an episode $\tau_i$ under $\pi_{\theta_i}$
for task $T_i$.
In general,
the goal of meta-RL is to train a meta-learner $f$,
which can quickly learn a policy (i.e., optimizing the parameter $\theta_i$) to minimize the negative expected reward for each task $T_i\in D$:
\begin{equation}
\label{eqn:loss_ti}
L_{T_i}(\tau_i|\theta_i)=-\mathbb{E}_{\tau_i\sim P_{i}(\tau_i|\theta_i)}[R(\tau_i)]
\end{equation}
where $R(\tau_i)=\sum_{t}r_{i,t}$.

Basically, the training procedure of meta-RL consists of two alternate stages: fast-update and meta-update.
During fast-update, the meta-learner
runs optimization several times (e.g., $3$ times) to obtain an adapted policy
for each task. During meta-update,
the meta-learner is optimized to
minimize the total loss of all tasks under the corresponding adapted policies.

\subsection{MAML}
Different meta-RL methods mainly differ in the design of
the meta-learner and fast-update.
Here, we will give a brief introduction
with MAML~\cite{finn2017model}
as an example.
The computation graph of MAML is shown in Figure~\ref{fig:cg_maml}.

MAML trains an SGD optimizer, parametrized by $\theta$, as the meta-learner.
During fast-update, for each task $T_i\in D$,
the meta-learner first initializes a policy network with $\theta_i^{1}=\theta$,
and then performs VPG update $K$ times.
The fast-update stage
is formulated as follows:
\begin{equation}
\theta_i^{k+1} = \theta_i^{k} - \alpha\nabla_{\theta_i^{k}} L_{T_i}(\tau_i^{k}|\theta_i^{k}), \quad k=1,\ldots,K
\end{equation}
where $\alpha$ is the learning rate and $K$ is the number of fast-updates.
Combined with meta-update,
MAML aims to learn a good policy initialization,
from which different parameters can be quickly learned for different tasks.

\section{Algorithm}
In this section, we propose a new meta-RL method TESP
that explicitly models the individuality and commonness of tasks.
Specifically,
TESP learns a shared policy to characterize the
task commonness, and simultaneously
trains a meta-learner to quickly abstract the individuality
to enable the shared policy to accomplish different tasks.
We will first introduce the overall network architecture of TESP,
and then elaborate how to leverage this architecture in a meta-learning manner.

\subsection{Network Architecture}

Here, we first introduce the network structure of TESP composed of
a task encoder and a policy network,
which is illustrated
in Figure~\ref{fig:arch}.

\subsubsection{Task Encoder}
The task encoder maps each task
into a low-dimensional latent space. It is expected that
the low-dimensional space $\mathbb{R}^d$ can capture
differences among tasks, such that we can represent the individuality of each
task using a low-dimensional vector $h\in \mathbb{R}^d$ named task embedding.
The first question is what kind of information we can use to learn such a
low-dimensional space.
In RL, an agent explores in an environment and generates
a bundle of episodes.
Obviously, these episodes contain characteristics of the ongoing task
which can be used to abstract some specific information about the task.

\begin{figure}[t]
\centering
\includegraphics[width=0.46\textwidth]{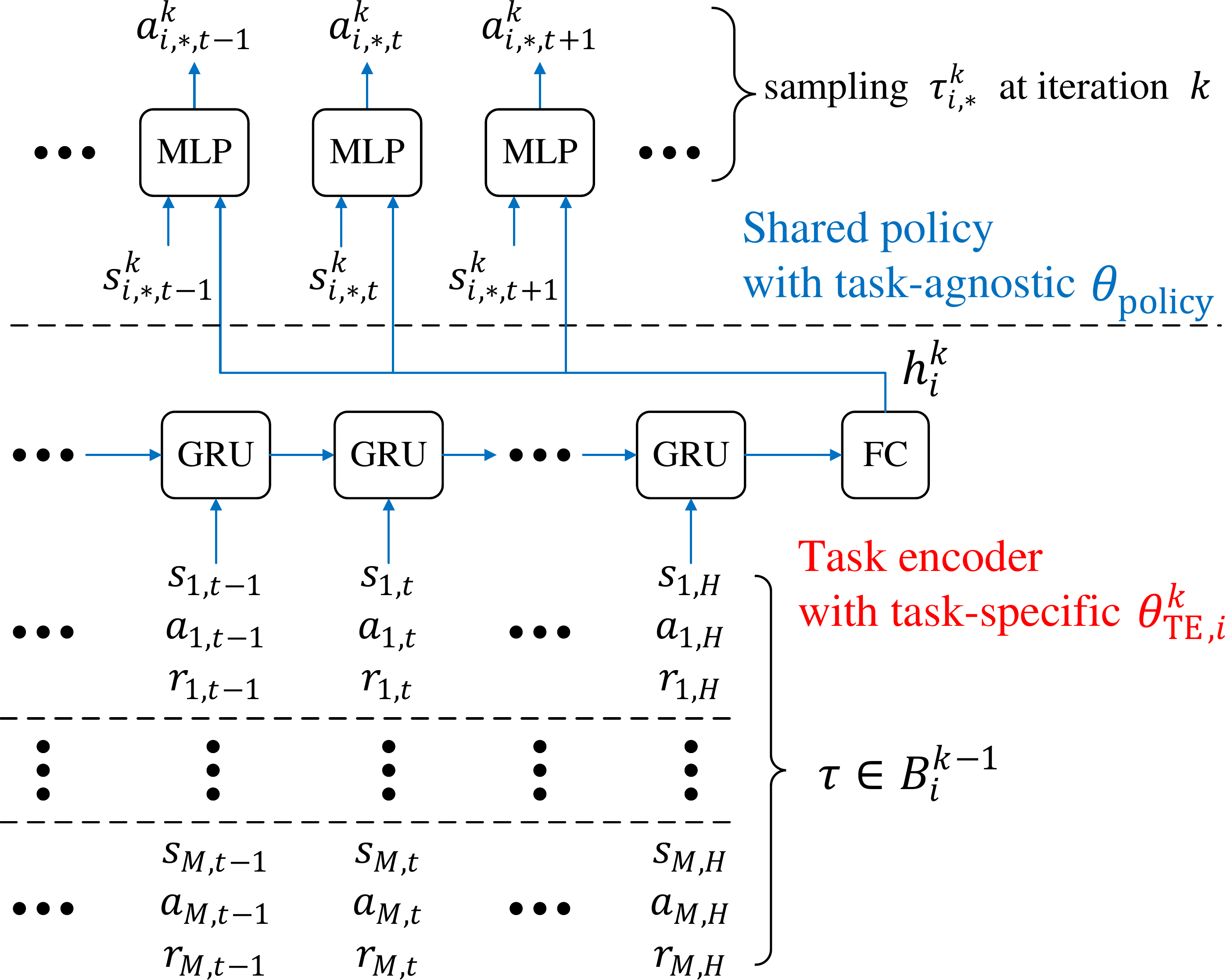}
\caption{
Network architecture of TESP
composed of a task encoder (an RNN with GRU cell followed by a fully-connected (FC) layer) and a policy network (an MLP network).
For each task $T_i$, when sampling episodes $\tau_{i,*}^{k}$ at the $k\nth$ iteration (or fast-update),
the task encoder first uses experienced episodes stored in the episode buffer $\tau\in B_i^{k-1}$
to calculate the corresponding task embedding $h_i^k$.
Then, the embedding $h_i^k$, along with the observation, is passed into the MLP network to
predict a distribution of actions at each timestep.
}
\label{fig:arch}
\end{figure}

Therefore, let us denote the task encoder by $g\colon E\to \mathbb{R}^d$,
where $E$ indicates the set of all episodes that an agent has experienced in an environment.
However,
simply using all episodes is computationally intensive in practice,
because we usually sample dozens of (e.g., 32) episodes at each iteration and the size of $E$
will increase rapidly.
Considering that our goal is to learn a discriminative embedding to characterize a task,
the episodes with low rewards are helpless
or even harmful as shown in \S~\ref{sec:ablation}.
To accelerate and boost
the learning process, we propose to build an episode buffer $B_i$
for each task $T_i\in D$,
which stores the best $M$ episodes an agent has experienced.
Mathematically, we initialize the buffer as an empty set $B_i^0=\varnothing$,
and then update the episode buffer as follows:
\begin{equation}
B_i^k = S_M(B_i^{k-1} \cup \{\tau_{i,1}^{k}, \ldots, \tau_{i,N}^{k}\}), \quad k=1,2,\ldots,
\end{equation}
where $B_i^k$ is the episode buffer after the $k\nth$ iteration\footnote{Hereafter, the \emph{iteration} means the \emph{fast-update} in the scope of meta-learning.},
$\tau_{i,\ast}^k$ is the episodes sampled at the $k^{\text{th}}$ iteration,
$N$ is the number of episodes sampled at each iteration,
and $S_M$ is a function that selects the best $M$ ($M<N$) episodes in terms of rewards:
\begin{equation}
S_M(X) = \argmax_{X'\subset X, |X'|=M} \sum_{\tau\in X'} R(\tau).
\end{equation}
Furthermore,
we use the episodes in the buffer to abstract the individuality of each task,
as shown in Figure~\ref{fig:arch}.
Mathematically, we have
\begin{equation}
\label{eqn:ta}
h_i^k = \frac{1}{M} \sum_{\tau\in B_i^{k-1}} g(\tau), \quad k=1,2,\ldots,
\end{equation}
where $g(\tau)$ refers to modeling an episode using the task encoder
and $h_i^{k}$ is the task embedding of task $T_i$
after the exploration of the $(k-1)^{\text{th}}$ iteration
(or before the exploration of the $k^{\text{th}}$ iteration).
Although a more principled way could be to
design a more comprehensive mechanism to
effectively and efficiently utilize all previously sampled episodes,
we empirically find that the simple episode buffer can achieve
good enough performance,
and we leave it as future work.

Given that an episode is a sequence of triplets $(s_t,a_t,r_t)$,
we model the task encoder $g$ as an RNN with GRU cell followed
by a fully-connected layer.
At each timestep, the GRU cell receives the concatenation of state, action, and reward
as shown in Figure~\ref{fig:arch}.

\subsubsection{Policy}
The policy $\pi$ predicts a distribution of actions based on the input.
Since we have modeled each task using the corresponding task embedding,
an agent can make decisions conditioned on the task-specific knowledge
in addition to raw observations.
Mathematically, we sample an episode
$\tau_i^k$
for task $T_i$
at the $k^{\text{th}}$ iteration:
\begin{equation}
\tau_{i}^{k} \sim \pi(\tau_i^k|h_i^{k}).
\end{equation}
At each timestep, the action is sampled as
\begin{equation}
\label{eqn:policy}
a_{i,t}^{k} \sim \pi(a_{i,t}^{k}|[s_{i,t}^{k},h_i^{k}]),
\end{equation}
where the input is the concatenation of the current observation and the
task embedding of $T_i$.
Note that $h_i^{k}$ represents the agent's understanding of task $T_i$,
and thus is appended to
each timestep of the sampling at the $k^{\text{th}}$ iteration.

\subsection{Meta-Learning}

As mentioned before, we aim to quickly learn some discriminative information (i.e., the individuality)
about a task, and model the commonness of a kind of tasks.
For the individuality, inspired by MAML~\cite{finn2017model},
we propose to train an SGD optimizer (i.e., the meta-learner) to quickly learn a task encoder $g_i$ for each task $T_i\in D$,
which further generates the corresponding task embedding.
For the commonness, we propose to learn a policy, which is shared across different tasks.
The shared policy is conditioned on task-specific embeddings,
which allows to accomplish different tasks with the same policy parameters.

While an alternative for the individuality
is to simply learn a single task encoder
and use the same set of parameters
to obtain task embeddings of different tasks based on the corresponding episode buffers,
we find that it
poorly generalizes to novel tasks
as shown in \S~\ref{sec:ablation}.

The parameters involved in TESP include
\begin{equation}
\theta = \{ \theta_{\text{TE}}, \alpha, \theta_{\text{policy}} \},
\end{equation}
where $\theta_{\text{TE}}$ and $\alpha$ are the initialization and the learning rate of the SGD optimizer,
and $\theta_{\text{policy}}$ is the parameter of the shared policy.
Empirically, we use adaptive per-parameter learning rates $\alpha$, which has been found
to have better performance than a fixed learning rate,
as in some prior works~\cite{li2017meta,al2017continuous,gupta2018meta}.

\subsubsection{Fast-update}
The purpose of the fast-update is to quickly optimize a task encoder $g_i$
for each task $T_i$ and obtain the corresponding task embedding,
which is formulated as
\begin{equation}
\label{eqn:tasp-sgd}
\theta_{\text{TE},i}^{k+1} = \theta_{\text{TE},i}^{k} - \alpha\circ
\nabla_{\theta_{\text{TE},i}^{k}} L_{T_i}(\tau_i^{k}|\theta_i^{k}), \quad k=1,\ldots,K.
\end{equation}
Here, $\theta_{\text{TE},i}^{1}=\theta_{\text{TE}}$,
$K$ is the number of fast-updates,
$\circ$ denotes Hadamard product,
and the definition of $L_{T_i}$ is the same as Eq.~(\ref{eqn:loss_ti}).
Due to that the episode buffer is empty at the beginning,
to make the derivation feasible at the first iteration,
we first warm up the episode buffer
by sampling a bundle of episodes $\tau_{i}^{\text{warm}}$
with the task embedding $h_i^{\text{warm}}$ assigned to a zero vector,
and then calculate $h_i^1$ and sample episodes $\tau_i^1$.

\begin{algorithm}[t]
\caption{Training Procedure of TESP}
\label{alg:meta-train}
{\small
\begin{algorithmic}[1]
\INPUT
training tasks $D$ and the number of fast-updates $K$
\OUTPUT the meta-learner $\theta_{\text{TE}}, \alpha$ and the shared policy $\theta_{\text{policy}}$
\STATE Randomly initialize $\theta_{\text{TE}}$, $\alpha$, and $\theta_{\text{policy}}$
\WHILE{\emph{not done}}
    \STATE Sample a batch of tasks $T_i\in D$
    \FORALL{$T_i$}
        \STATE Initialize $\theta_i^{1}=\{\theta_{\text{TE}}, \alpha, \theta_{\text{policy}}\}$
        \STATE Sample episodes $\tau_{i}^{\text{warm}}$ with $h_i^{\text{warm}}=\vec{0}$,
        and warm up $B_i^0 = S_M(\{\tau_{i,1}^{\text{warm}}, \ldots, \tau_{i,N}^{\text{warm}}\})$
        \FOR{$k\in\{1,\ldots,K\}$}
            \STATE
            Calculate \textbf{task embedding} $h_i^{k}$ via Eq.~(\ref{eqn:ta}) using $\theta_{\text{TE},i}^{k}$
            \STATE
            Sample episodes $\tau_i^{k}$ using $h_i^{k}$ and $\theta_{\text{policy}}$
            \STATE
            Perform \textbf{fast-update} \\ $\theta_{\text{TE},i}^{k+1}=\theta_{\text{TE},i}^{k}-\alpha\circ\nabla_{\theta_{\text{TE},i}^{k}} L_{T_i}(\tau_i^{k}|\theta_i^{k})$
            \STATE
            Update the \textbf{episode buffer} \\
            $B_i^k = S_M (B_i^{k-1}\cup\{\tau_{i,1}^{k}, \ldots, \tau_{i,N}^{k}\})$
        \ENDFOR
        \STATE
        Calculate \textbf{task embedding} $h_i^{K+1}$, and
        sample episodes $\tau_i^{K+1}$ using $h_i^{K+1}$ and $\theta_{\text{policy}}$
    \ENDFOR
    \STATE
    Update $\theta_{\text{TE}}$, $\alpha$, and $\theta_{\text{policy}}$
    to optimize the objective function~(\ref{eqn:obj})
\ENDWHILE
\end{algorithmic}
}
\end{algorithm}

\subsubsection{Meta-update}
During meta-update, we optimize the parameters of the SGD optimizer and the policy together
to minimize the following objective function:
\begin{equation}
\label{eqn:obj}
\resizebox{0.99\hsize}{!}{$
\displaystyle
\min_{\theta_{\text{TE}}, \alpha, \theta_{\text{policy}}}
\sum_{T_i\in D}L_{T_i}(\tau_i^{K+1}|\theta_{\text{TE},i}^{K+1},\alpha,\theta_{\text{policy}})
+ \eta \sum_{i} \|h_i^{K+1}\|^2,
$}
\end{equation}
where $\eta>0$ is a constant factor that balances the effects of the two terms.
Here, we propose to improve the generalization ability from two aspects:
(1)
The parameter $\theta_{\text{policy}}$ is only optimized w.r.t. all tasks during meta-update
(without adaptation during fast-update),
which enforces that a versatile policy is learned;
(2)
The second term in Eq.~(\ref{eqn:obj}) acts as a regularizer to constrain that
task embeddings of different tasks are not so far from the origin point such that
the shared policy cannot learn to cheat.
This term is inspired by VAE~\cite{kingma2013auto},
where the KL divergence between the learned distribution and a normal distribution
should be small.
We perform ablation studies on these two aspects in \S~\ref{sec:ablation}.
A concise training procedure is provided in Algorithm~\ref{alg:meta-train}.

\subsubsection{Adaptation to Novel Tasks}
At testing time, we have a set of novel tasks,
and expect to learn
these tasks as efficiently as possible.
We have obtained an SGD optimizer and a shared policy.
The SGD optimizer is able to quickly learn a task encoder to
abstract the individuality of a task represented by a task embedding,
and the shared policy is able to accomplish different tasks
conditioned on different task embeddings.
Therefore, for each novel task,
we simply sample episodes and employ the SGD optimizer to learn a task encoder
to acquire the appropriate task embedding $h_i$ according to
Eq.~(\ref{eqn:ta})~and~(\ref{eqn:tasp-sgd}),
while the policy does not need further adaptation.

\section{Experiments}

\begin{figure}[t]
\centering
\begin{subfigure}[b]{0.1\textwidth}
\includegraphics[width=\textwidth]{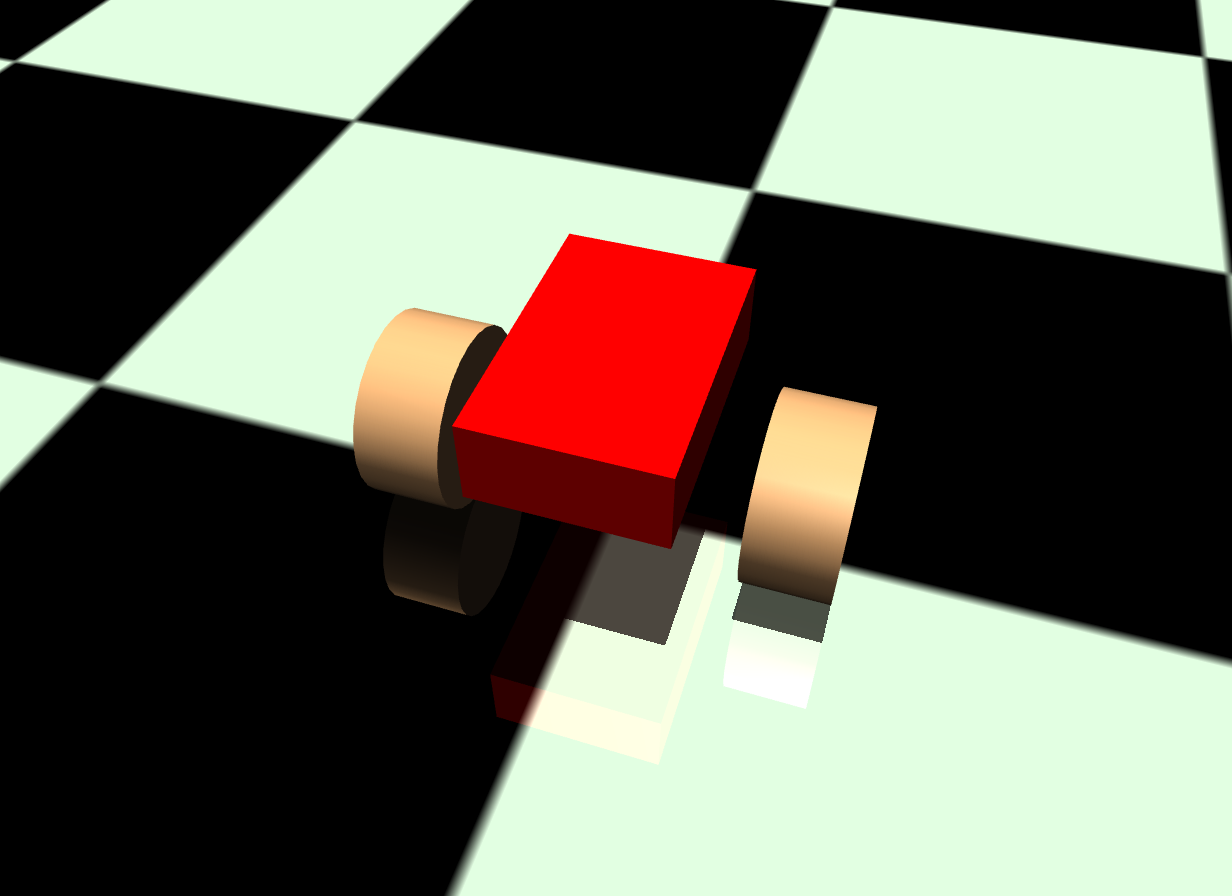}
\caption{}
\label{task:wheeled}
\end{subfigure}
\hspace{3pt}
\begin{subfigure}[b]{0.1\textwidth}
\includegraphics[width=\textwidth]{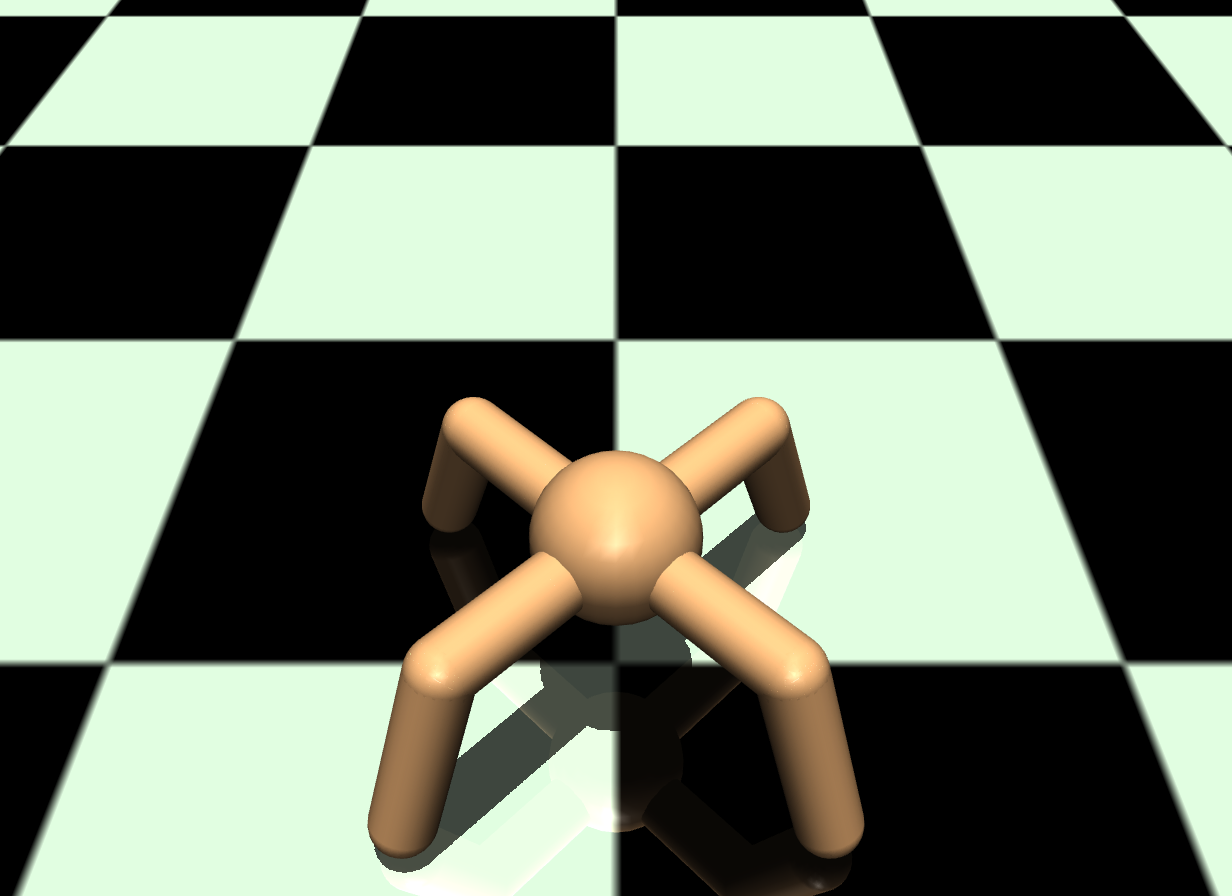}
\caption{}
\label{task:ant}
\end{subfigure}
\hspace{3pt}
\begin{subfigure}[b]{0.1\textwidth}
\includegraphics[width=\textwidth]{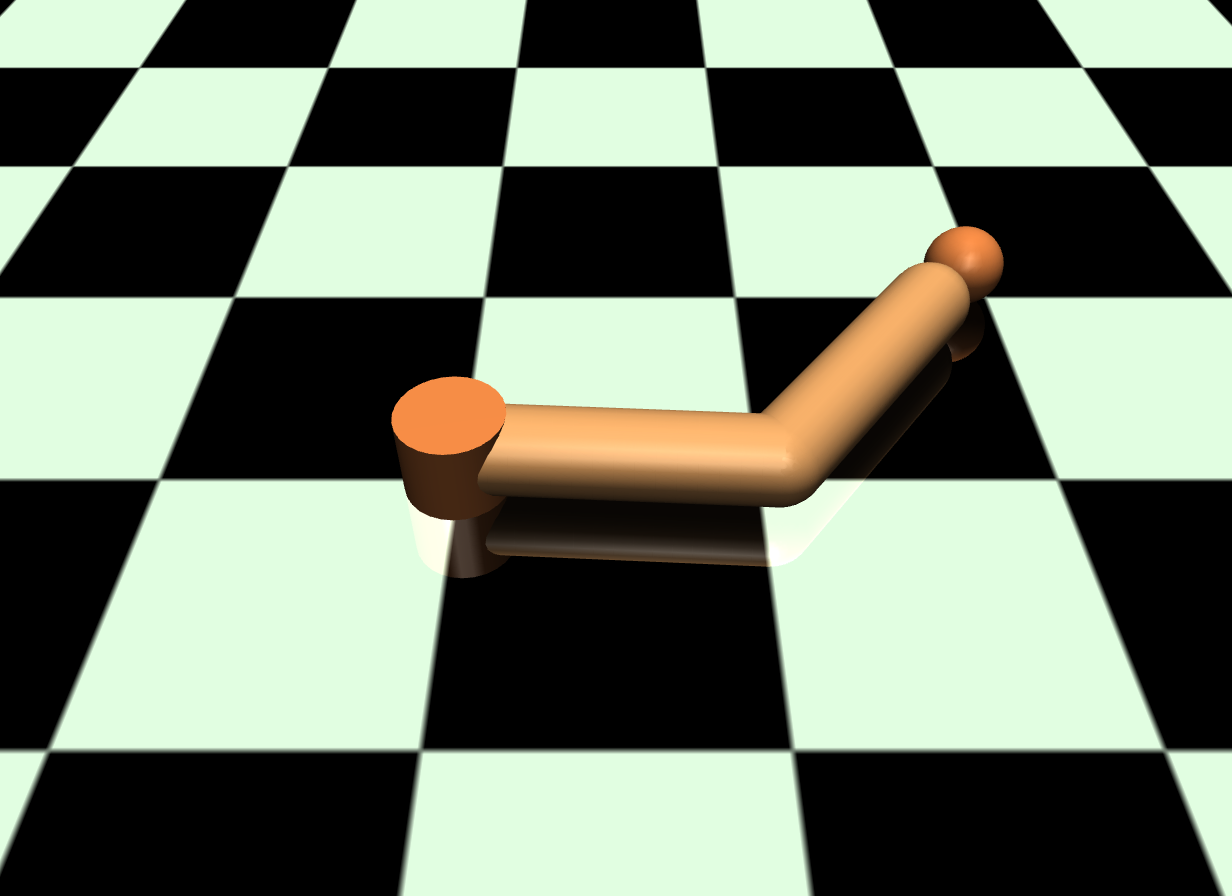}
\caption{}
\label{task:2-link}
\end{subfigure}
\hspace{3pt}
\begin{subfigure}[b]{0.1\textwidth}
\includegraphics[width=\textwidth]{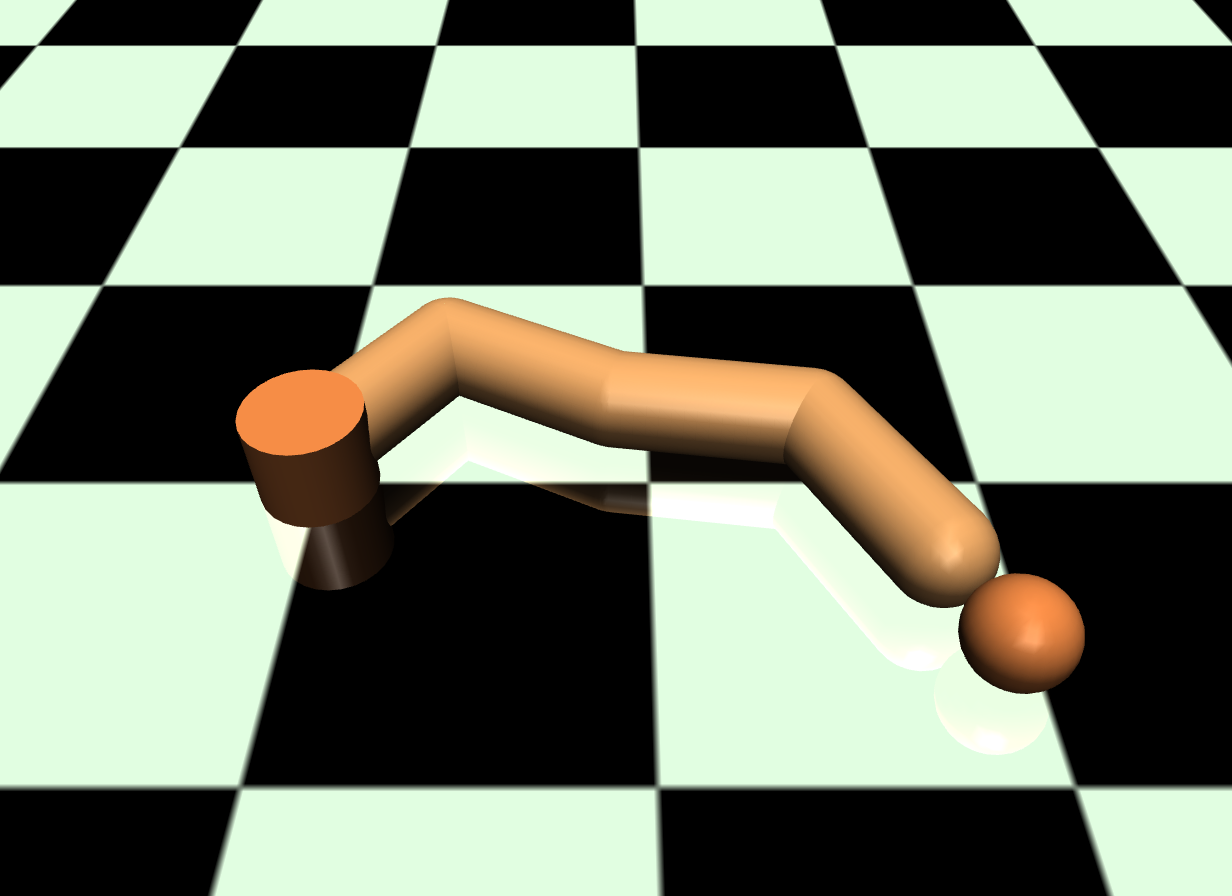}
\caption{}
\label{task:4-link}
\end{subfigure}
\caption{
Renderings of the
(\subref{task:wheeled}) wheeled agent,
(\subref{task:ant}) quadrupedal ant,
(\subref{task:2-link}) 2-link reacher,
and
(\subref{task:4-link}) 4-link reacher.
}
\label{fig:tasks}
\end{figure}

In this section,
we comparatively evaluate our proposed method
on four tasks
with MuJoCo simulator~\cite{todorov2012mujoco}:
(1) a wheeled agent attempting to reach different target locations,
(2) a quadrupedal ant attempting to reach different target locations,
(3) a 2-link reacher attempting to reach different end-effector target locations,
(4) a 4-link reacher attempting to reach different end-effector target locations.
Figure~\ref{fig:tasks} shows the renderings of agents used in the above tasks.

\subsection{Experimental Settings}

For each family of tasks, we sample $100$ target locations
within a circle $\rho<R_1$
as training tasks $D$.
When it comes to testing,
we consider two scenarios:
(1) Sample another $100$ target locations
within the circle $\rho<R_1$
as novel/testing tasks $D'$ (i.e., from the same distribution);
(2) Sample $100$ target locations
within an annulus $R_1<\rho<R_2$
as novel tasks $D''$ (i.e., from a different distribution).
The wheeled and ant agents always start from the origin point,
and the reachers are placed along the horizontal direction at the beginning.

We compare TESP with three baselines:
MAML~\cite{finn2017model},
Meta-SGD~\cite{li2017meta},
and TESP with a single variable being optimized during fast-update (denoted by AdaptSV) analogously to MAESN~\cite{gupta2018meta}.
Here, we did not consider recurrent meta-learners
such as RL$^2$~\cite{duan2016rl} and SNAIL~\cite{mishra2017meta},
due to that prior works have shown that recurrent meta-learners tend to
completely overfit to the distribution of training tasks and
cannot generalize to out-of-distribution tasks
(i.e., $D''$ in this paper).
We did not include some traditional HRL baselines that have a similar architecture to our method,
because they are generally not suitable to our scenarios
where we consider fast learning on novel tasks.
For example,
\cite{florensa2017stochastic} requires training an extra neural network from scratch when learning a novel task,
which is almost impossible to converge in $3$ iterations.

For each method, we set the number of fast-updates $K$ to $3$,
and use the first-order approximation during fast-update to speed up
the learning process as mentioned in~\cite{finn2017model}.
We use VPG~\cite{williams1992simple}
to perform fast-update, and PPO~\cite{schulman2017proximal}
to perform meta-update.
For detailed settings of environments and experiments,
please refer to
the supplement
at~\url{\codeurl}.

\begin{figure}[!t]
\centering
\begin{subfigure}[b]{0.45\textwidth}
\includegraphics[width=\textwidth]{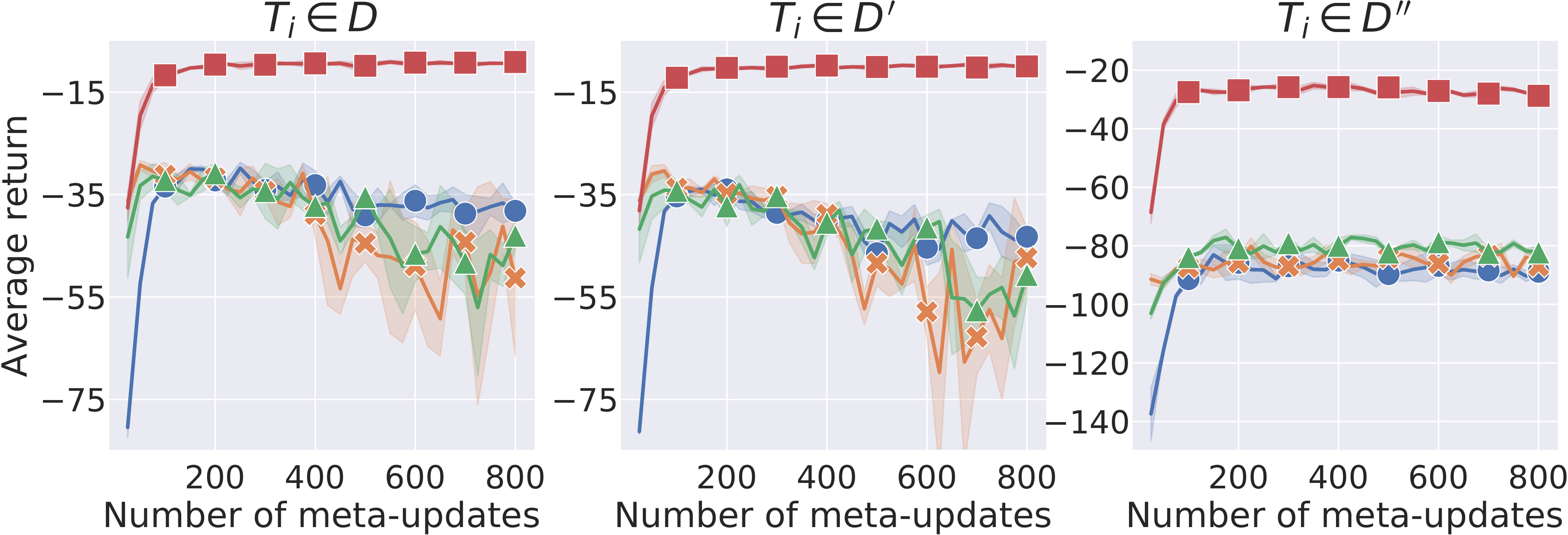}
\caption{Wheeled locomotion.}
\label{return:wheeled}
\end{subfigure}
\begin{subfigure}[b]{0.45\textwidth}
\vspace{3pt}
\includegraphics[width=\textwidth]{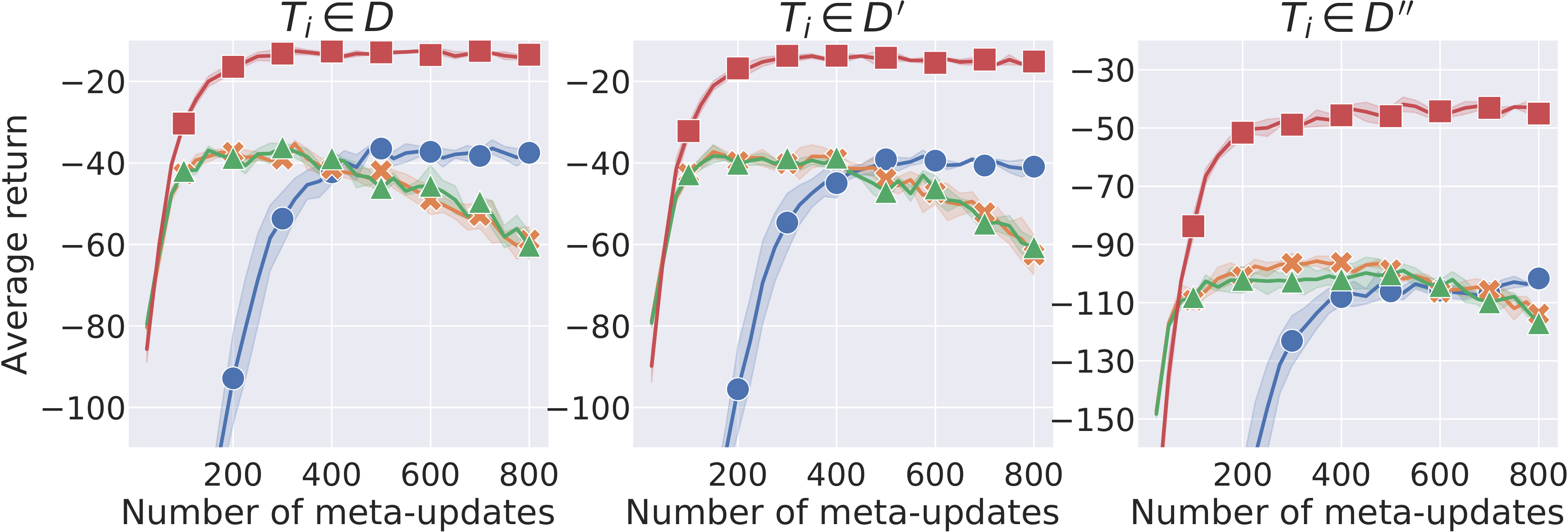}
\caption{Ant locomotion.}
\label{return:ant}
\end{subfigure}
\begin{subfigure}[b]{0.45\textwidth}
\vspace{3pt}
\includegraphics[width=\textwidth]{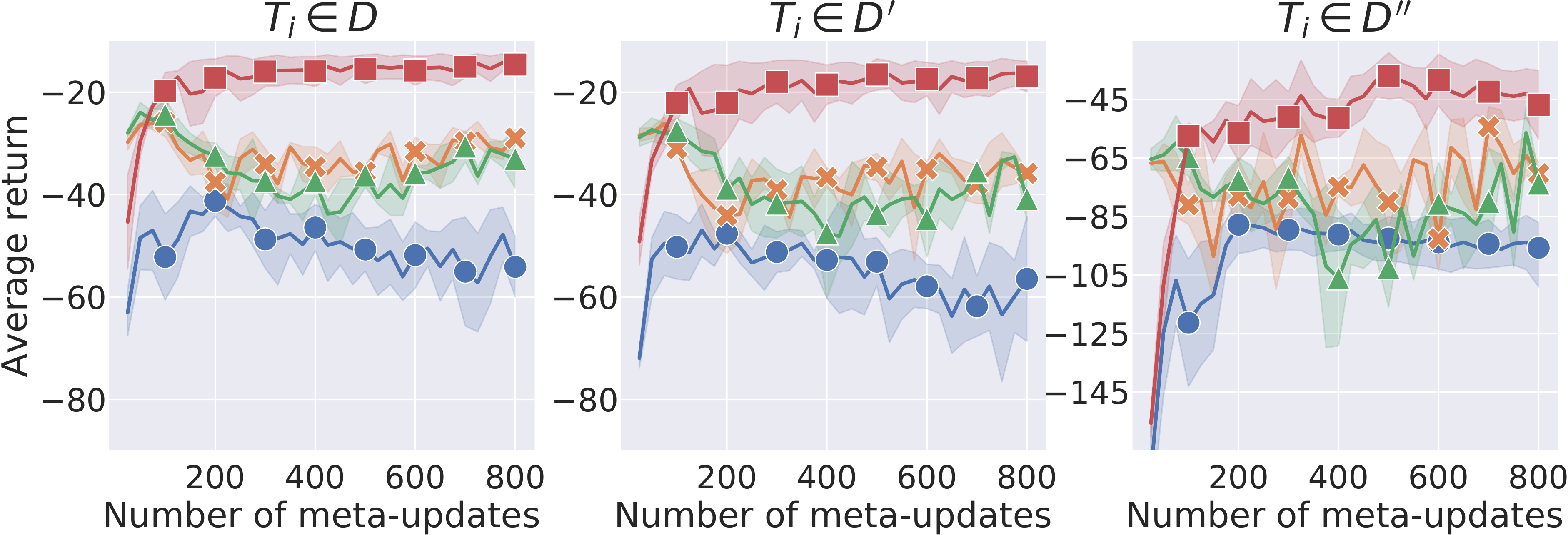}
\caption{2-link reacher.}
\label{return:2-link}
\end{subfigure}
\begin{subfigure}[b]{0.45\textwidth}
\vspace{3pt}
\includegraphics[width=\textwidth]{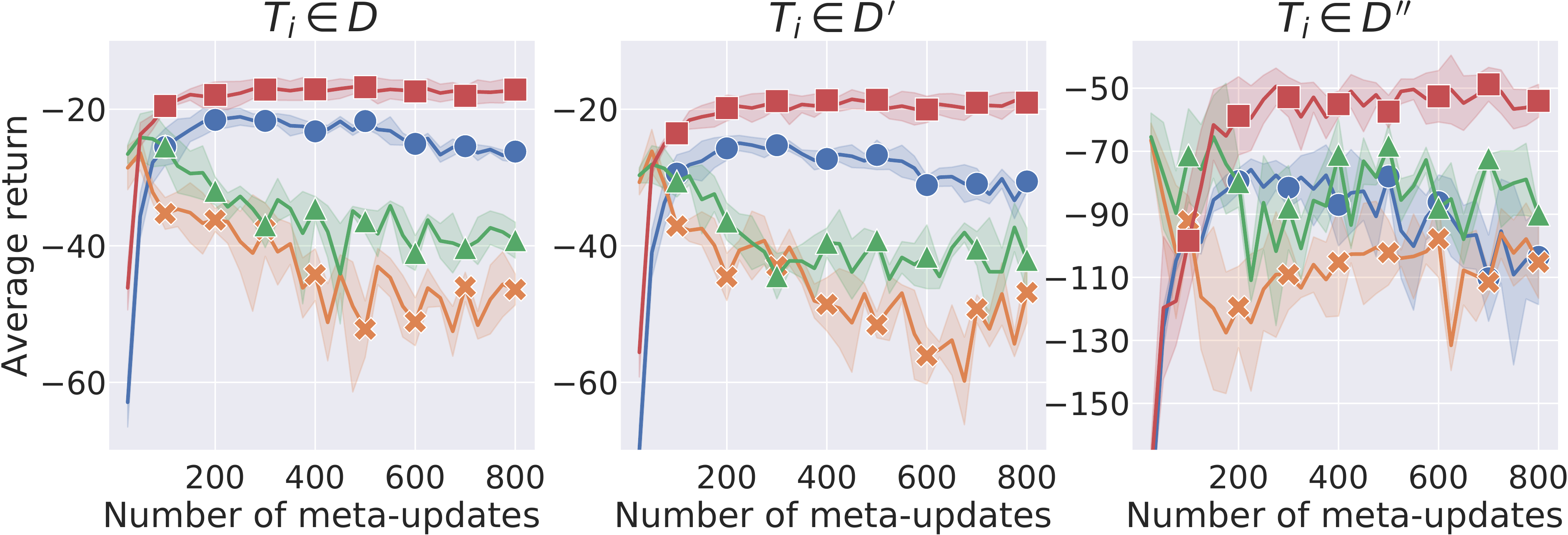}
\caption{4-link reacher.}
\label{return:4-link}
\end{subfigure}
\begin{subfigure}[b]{0.35\textwidth}
\vspace{3.0pt}
\includegraphics[width=\textwidth]{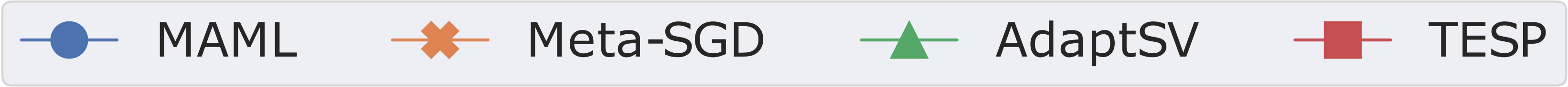}
\end{subfigure}
\caption{
Learning curves on training and testing tasks.
The y-axis represents the average return after $3$ fast-updates over all tasks.
We directly apply the models trained on $D$ to perform evaluations,
and thus testing tasks $D'$ and $D''$ are not seen during the training of models.
The plots are averaged over the best performing $3$ random seeds out of $10$ random seeds.
For easy reading,
the y-axes of some plots are truncated
and the markers are plotted every $100$ meta-updates.
}
\label{fig:returns}
\end{figure}

\subsection{Empirical Results}
To better reflect the learning ability on training tasks
and the generalization ability on novel tasks,
we plot learning curves of different methods on both training and novel tasks
as shown in Figure~\ref{fig:returns}.
Specifically,
we perform evaluations on $D$, $D'$, and $D''$ every $25$ meta-updates.
In each evaluation, we apply the models of different methods to perform $3$ fast-updates
for each task, and report the average return after fast-updates over all tasks.
The reported returns are calculated by
$R = \frac{1}{100} \sum_{T_i} \sum_{t} r_{i,t}^{\text{dist}}$,
where $100$ indicates the size of $D$, $D'$, or $D''$, and
$r_{i,t}^{\text{dist}}$ is the distance reward
which is the negative distance to the target location.

From Figure~\ref{fig:returns}, we observe that
TESP significantly outperforms all baselines on $D$,
which indicates TESP has better learning capacity than baselines on training tasks
and is expected since TESP uses a more complex network
(i.e., an extra RNN for the task encoder).
In addition,
all methods including our TESP and baselines
have similar learning curves on $D$ and $D'$,
which demonstrates their ability to generalize to
novel tasks sampled from the training distribution.
However,
the baselines tend to overfit to the training distribution
and show poor performance on out-of-distribution tasks $D''$,
but our TESP still has good performance on $D''$.
Moreover,
the gap between the performance of TESP on training
and out-of-distribution tasks is smaller than those of baselines.
Therefore, the reason why TESP shows better performance on $D''$
is not only that TESP learns training tasks better,
but also that TESP is more generalizable.

The comparison with AdaptSV shows that
simply adapting a single variable is not enough to represent different tasks.
In contrast, our method is able to efficiently obtain
a task embedding to represent each task by leveraging past experience
stored in an episode buffer with a meta-learned task encoder.
On the other hand,
the convergence of TESP is more stable as the number of meta-updates increases,
and the variance of TESP over different random seeds is smaller than baselines.

\subsection{Ablation Studies}
\label{sec:ablation}
\begin{figure}[t]
\centering
\begin{subfigure}[b]{0.48\textwidth} %
\includegraphics[width=\textwidth]{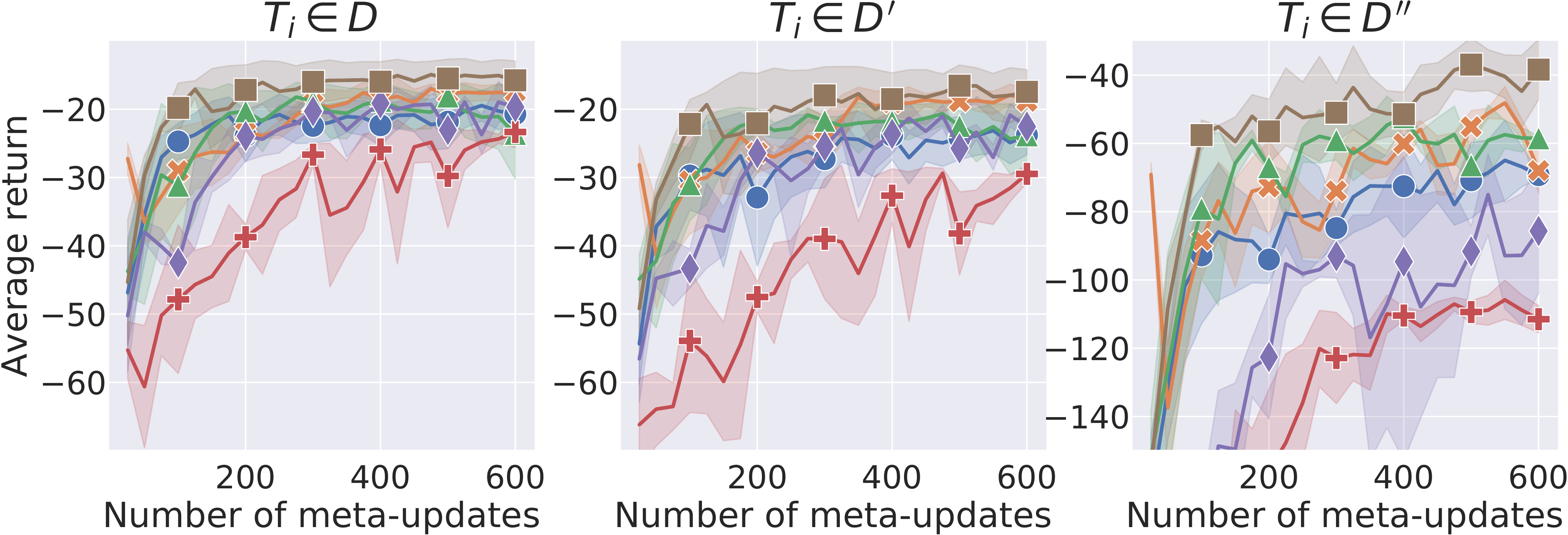}
\end{subfigure}
\begin{subfigure}[b]{0.38\textwidth} %
\vspace{5pt}
\includegraphics[width=\textwidth]{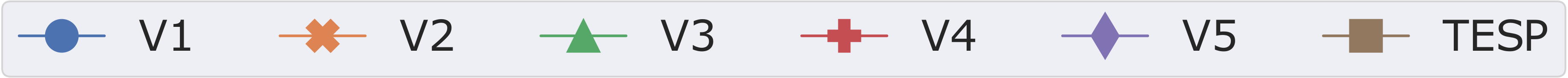}
\end{subfigure}
\caption{
Ablation studies on 2-link reacher tasks.
}
\label{fig:ablation}
\end{figure}

Since we introduce several different ideas into TESP,
including
the episode buffer holding the best $M$ experienced episodes for each task,
the learnable SGD optimizer for task encoders,
the shared policy,
the regularization term in Eq.~(\ref{eqn:obj}),
and
adaptive per-parameter learning rates of the learnable SGD optimizer,
we perform ablations to investigate the contributions of these different ideas.
Variants considered are
(1) the episode buffer holding all experienced episodes for each task,
(2) additionally fast-updating the policy for each task,
(3) $\eta=0$ (i.e., without the regularization term),
(4) $\alpha=0$ (i.e., without the SGD optimizer for fast-updating task encoders),
and
(5) holding constant the learning rate of the SGD optimizer.
From Figure~\ref{fig:ablation},
we observe that most variants have
similar performance to TESP on $D$ and $D'$,
but perform much worse on $D''$.
The comparison with V1 shows that
episodes with low rewards have a bad impact on
the learning of task embeddings.
Comparing TESP with V2 and V3,
we confirm that the shared policy and the regularization term
enable better generalization, especially for out-of-distribution novel tasks.
The results of V4 indicate that it is crucial to leverage the proposed architecture
in a meta-learning manner.
As in prior works~\cite{li2017meta,al2017continuous,gupta2018meta},
we also find that adaptive per-parameter learning rates
can lead to better performance
by comparing TESP with V5.

\section{Conclusion}
In this work,
we presented TESP,
of which the basic idea is
to explicitly model the individuality and commonness of tasks in the scope of meta-RL.
Specifically,
TESP trains a shared policy and an SGD optimizer coupled to a task encoder network
from a set of tasks.
When it comes to a novel task, we apply the SGD optimizer to quickly learn a task encoder
which generates the corresponding task embedding,
while the shared policy remains unchanged
and just predicts actions based on observations and the task embedding.
In future work,
an interesting idea would be to
extend TESP with a set of shared conditional policies inspired by~\cite{frans2017meta}.

\section*{Acknowledgments}
We gratefully thank Fei Chen and George Trimponias for insightful discussions
and feedback on early drafts.
The research presented in this paper is supported in part by National Key R\&D Program of China (2018YFC0830500), National Natural Science Foundation of China (U1736205, 61603290), Shenzhen Basic Research Grant (JCYJ20170816100819428), Natural Science Basic Research Plan in Shaanxi Province of China (2019JM-159), and Natural Science Basic Research in Zhejiang Province of China (LGG18F020016).

\bibliographystyle{named}
\bibliography{references}
\end{document}